\documentclass[twocolumn,english,journal,letterpaper,oneside,twocolumn]{IEEEtran}
\usepackage[T1]{fontenc}
\usepackage{babel}
\usepackage{array}
\usepackage{verbatim}
\usepackage{float}
\usepackage{multirow}
\usepackage{amsmath}
\usepackage{amssymb}
\usepackage{graphicx}
\usepackage{xcolor}
\usepackage{color}
\usepackage{gensymb}
\usepackage[unicode=true,
 bookmarks=false,
 breaklinks=false,pdfborder={0 0 0},backref=false,colorlinks=false]
 {hyperref}
 
\usepackage{caption}

\makeatletter

\providecommand{\tabularnewline}{\\}
\floatstyle{ruled}
\newfloat{algorithm}{tbp}{loa}
\providecommand{\algorithmname}{Algorithm}
\floatname{algorithm}{\proV1tect\algorithmname}

\@ifundefined{date}{}{\date{}}
\makeatother

\begin{document}

\title{Semi-supervised Large-scale Fiber Detection in Material Images with Synthetic Data}

\author{Lan Fu, \IEEEmembership{Student Member, IEEE}, Zhiyuan Liu, Jinlong Li, Jeff Simmons, \IEEEmembership{Senior Member, IEEE}, Hongkai Yu\IEEEauthorrefmark{1}, \IEEEmembership{Member, IEEE}, and Song Wang\IEEEauthorrefmark{1}, \IEEEmembership{Senior Member, IEEE}%
\thanks{Lan Fu and Song Wang are with the Department of
Computer Science \& Engineering, University of South Carolina, SC, USA, e-mail: \protect\href{mailto:lanf@email.sc.edu}{lanf@email.sc.edu}, 
\protect\href{mailto:songwang@cec.sc.edu}{songwang@cec.sc.edu}.
}%
\thanks{Zhiyuan Liu is with the Department of Computer Science, University of North Carolina at Chapel Hill, NC,
USA, e-mail: \protect\href{mailto:zhiy@cs.unc.edu}{zhiy@cs.unc.edu}.
}%
\thanks{Jinlong Li is with the School of Information Engineering, Chang'an University, Xi'an,
China, e-mail: \protect\href{mailto:lijinlong1117@chd.edu.cn}{lijinlong1117@chd.edu.cn}.
}%
\thanks{Hongkai Yu is with the Department of
Electrical Engineering and Computer Science, Cleveland State University, OH,
USA, e-mail: \protect\href{mailto:h.yu19@csuohio.edu}{h.yu19@csuohio.edu}.
}%
\thanks{Jeff Simmons is with the Materials and Manufacturing Directorate, Air Force Research Lab, OH, USA, e-mail: \protect\href{mailto:jeff.simmons.3@us.af.mil}{jeff.simmons.3@us.af.mil}.
}%
\thanks{* indicates the  co-corresponding authors: Hongkai Yu, Song Wang}.
}%
\maketitle

\begin{abstract}
Accurate detection of large-scale, elliptical-shape fibers, including their parameters of center, orientation and major/minor axes, on the 2D cross-sectioned image slices is very important for characterizing the underlying cylinder 3D structures in microscopic material images. Detecting fibers in a degraded image poses a challenge to both current fiber detection and ellipse detection methods. This paper proposes a new semi-supervised deep learning method for large-scale elliptical fiber detection with synthetic data, which frees people from heavy data annotations and is robust to various kinds of image degradations. A domain adaptation strategy is utilized to reduce the domain distribution discrepancy between the synthetic data and the real data, and a new Region of Interest (RoI)-ellipse learning and a novel RoI ranking with the symmetry constraint are embedded in the proposed method. Experiments on real microscopic material images demonstrate the effectiveness of the proposed approach in large-scale fiber detection.  


\end{abstract}

\section{Introduction\label{sec:Introduction}}

\begin{figure}[htbh]
\centering
\includegraphics[clip=true, width=0.5\textwidth]{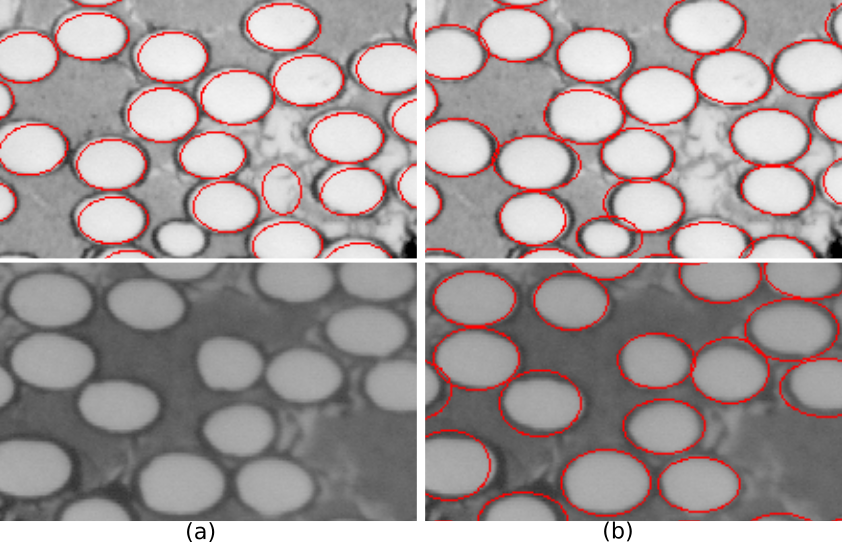} 
\caption{Large-scale elliptical fiber detection by (a) traditional Hough transform based ellipse fitting method EMMPMH~\cite{zhou2016large} and (b) the proposed method on clean (top) and degraded (bottom) images.}
\label{fig:ellipse-vis}
\end{figure}
\IEEEPARstart{C}{ontinuous} fiber reinforced composite (FRC) materials play a critical role in automotive and aerospace industries~\cite{knox2001continuous,suresh2013fundamentals}, due to notably better strength and stiffness than traditional materials. Accurate characterization of fiber's underlying microstructure, determining those superior properties, can substantially facilitate new material design and development. Extracting and characterizing large-scale (curved) cylinder structures of fibers are desired in microscopic material images. The first step to achieve this goal is to accurately detect them on each 2D cross-sectioned image slice. Given non-perpendicular sectioning planes, the cross-section of such a cylinder is usually an elliptical region, instead of a perfect circular region, as shown in Fig.~\ref{fig:ellipse-vis}. In practice, both location and shape of these ellipses are important for further characterization of the underlying 3D cylinder structures. For example, from the orientation and semi-major/semi-minor axis information of the detected elliptical fiber, we can infer the direction of the underlying 3D cylinder. In this paper, we will address the elliptical fiber detection by estimating all five key parameters $[x,y, \theta, R, r]$, where $(x,y)$ is the center location of the ellipse, $\theta$ is the orientation angle, i.e., the angle between the major axis and the horizontal axis, $R$ is the semi-major axis and $r$ is the semi-minor axis. 

In real-world applications, the proposed elliptical fiber detection can be quite challenging due to geometry variations of ellipse parameters and various kinds of image degradations, such as stains and blurs. Degradation effects might exist in local image regions due to contaminants during manufacturing  process~\cite{yu2018unsupervised}. 
Current elliptical fiber detector (EMMPMH)~\cite{zhou2016large} segments fiber regions first by a MRF-based labeling  method~\cite{comer2000mpm} and then a Hough transform~\cite{xie2002new} based ellipse fitting algorithm is utilized to obtain the five key parameters of fibers.
Many traditional ellipse detection methods, like Hough transform based methods~\cite{xie2002new,lu2008detection,tang2011ellipse,chia2007ellipse} and edge following based methods~\cite{puatruaucean2016jointa,fornaciari2014fast,jia2017fast,fan2015fiducial,lu2019arc,meng2020arc}, detect ellipses based on edge information with large computation costs. However, they may fail to detect large-scale elliptical fibers due to missing and inaccurate edge pixels in the above-mentioned degraded situations. Our main idea is to use the advanced deep learning method to improve the elliptical fiber detection in both clean and degraded image regions.

Recently, based on the Convolutional Neural Networks (CNNs), many object detectors, e.g., Faster R-CNN~\cite{ren2015faster}, SSD~\cite{liu2016ssd}, YOLO~\cite{redmon2016you}, achieved promising detection performances in computer vision tasks. Using Faster R-CNN and the spatio-temporal consistency in object tracking, a deep learning method~\cite{yu2018unsupervised} is proposed for fiber detection. These CNN-based deep learning methods have two drawbacks that prevent their application to the proposed large-scale elliptical fiber detection: 1) These CNN-based methods require a large number of manually annotated data as training set for CNN training. In practice, manually annotating the five key parameters for large-scale elliptical objects is very hard and time-consuming. 2) These CNN-based methods mainly detect the bounding box for each object~\cite{ren2015faster,liu2016ssd,redmon2016you,yu2016groupwise, he2017mask,yu2018unsupervised}, but not the desired five key parameters of the ellipse. In order to overcome these two drawbacks of CNN-based deep learning methods, we use synthetic ellipse data to reduce the workload of the manual annotations and design a new Region of Interest (RoI)-ellipse learning method to estimate the five key ellipse parameters.

In this paper, we propose a new semi-supervised deep learning method for large-scale elliptical fiber detection with the help of synthetic ellipse data, which is robust to various kinds of image degradations. 
We use a domain adaptation strategy to reduce the domain distribution discrepancy between the synthetic data and the real data. Inspired by Ding \textit{et al}.~\cite{ding2019learning}, a new RoI-ellipse learning is proposed to estimate the five ellipse parameters. Furthermore, considering the symmetry of ellipse, a novel RoI ranking with the symmetry constraint is incorporated into the proposed method. In summary, the main contributions of this paper are:
 
\begin{itemize}
\item {We propose a new semi-supervised learning method which trains on the synthetic data and tests on the real data to reduce the heavy burden of manual annotations, where CycleGAN~\cite{zhu2017unpaired} is used for 
domain adaptation to reduce the domain distribution 
discrepancy between the synthetic and the real
data.}

\item By designing a new RoI-ellipse learning, the proposed method is able to estimate the five key parameters of elliptical fibers on both clean and degraded images.

\item A new RoI ranking method with symmetry constraint is proposed to improve the detection accuracy.   
\end{itemize}

\section{Related Work\label{sec:Related-Work}}
\noindent \textbf{Elliptical object detection.} For elliptical-shaped fiber detection in material images, Zhou \textit{et al}.~\cite{zhou2016large} combines EMMPM segmentation algorithm~\cite{comer2000mpm} to extract fiber regions followed by a Hough-transform (HT) based ellipse fitting~\cite{xie2002new}. However, it fails when there are various kinds of degradations in the image. Traditional ellipse detectors include HT based algorithms and edge following based methods. HT~\cite{xie2002new} can map edge pixels of ellipse into a 5-dimensional parameter space. This mapping is vulnerable to substantial image noises and complicated backgrounds and computationally expensive in practice. 
Randomized Hough Transform (RHT)~\cite{xu1990new} obtained the ellipse parameters by mapping five random-chosen edge pixels into a point of 5-dimensional ellipse parameter space.
Mclaughlin \textit{et al}.~\cite{mclaughlin1998randomized} improved the RHT performance by a two-stage decomposition method. Chia \textit{et al}.~\cite{chia2007ellipse} presented a novel ellipse detection algorithm based on HT utilizing a one-dimensional accumulator to optimize the memory storage and computational complexity. Lu and Tan~\cite{lu2008detection} proposed to iteratively apply the RHT to a region of interest in the image space in order to remove noisy pixels. Constraint Random Hough Transform (CRHT)~\cite{tang2011ellipse} exploited constraints on sampled points to select high quality ones for ellipse fitting. The edge following based methods detect the arc segments from image to exploit abundant gradient and geometric constraints. The parameterless Ellipse and Line Segment Detector (ELSD)~\cite{puatruaucean2016jointa} combined line segment and elliptical arc detector based on the number of false alarm (NFA) method. Fornaciar \textit{et al}.~\cite{fornaciari2014fast} segmented edge contours into four classes based on arc-level and determined the convexity of the arcs with at least three arc segments by an innovative arc selection strategy. Jia \textit{et al}.~\cite{jia2017fast} proposed an ellipse detection method by pruning and selecting candidates using a modified projective invariant~\cite{fan2015fiducial} based on characteristic number. Lu \textit{et al}.~\cite{lu2019arc} proposed an arc-support line segment based ellipse detector in an accurate and efficient way. The consecutive arc-support line segments, sharing similar geometric properties, are iteratively linked to form arc-support groups. Meng \textit{et al}.~\cite{meng2020arc} proposed a fast and accurate ellipse detection algorithm utilizing digraph-based arc adjacency matrix and bidirectional search strategy (AAMED). The aforementioned elliptical object detection methods are generally prone to noise and complicated background, usually not robust enough to detect fibers in degraded image regions. In this paper, we develop advanced CNN-based object detector to improve the robustness of fiber detection.

\begin{figure*}[htbh]
\centering
\includegraphics[width=0.88\textwidth]{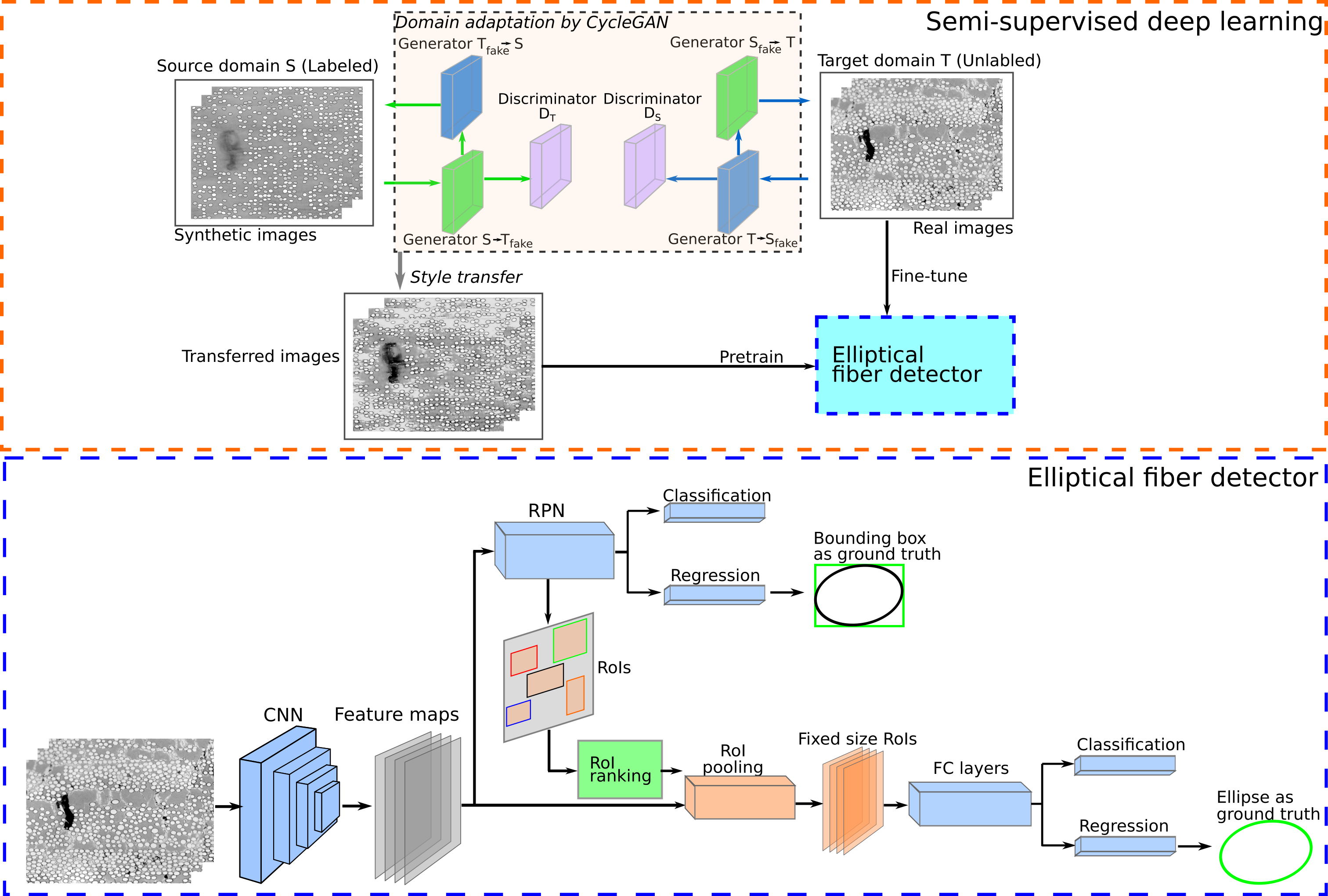} 
\caption{Framework of the proposed semi-supervised deep learning method for large-scale elliptical fiber detection.}
\label{fig:semi}
\end{figure*}

\textbf{CNN-based object detectors.} Many CNN-based deep object detectors, including one-stage (SSD~\cite{liu2016ssd}, YOLO~\cite{redmon2016you}) and two-stage (Faster R-CNN~\cite{ren2015faster}, Mask R-CNN~\cite{he2017mask}) detection pipelines, achieved promising detection performance in real-world applications. However, most of these CNN-based methods require a large number of manually annotated data for CNN training. CNN-based methods mainly detect bounding boxes for each object~\cite{ren2015faster,liu2016ssd,redmon2016you,yu2016groupwise, he2017mask,yu2018unsupervised}, which is not sufficient to infer five key ellipse parameters. In this paper, we propose a semi-supervised large-scale elliptical fiber detection framework by utilizing synthetic data and a small set of manually annotated real data. Because there are large-scale fibers with different scales and arbitrary orientations in cluttered arrangement on each image slice, we build our detection framework on two-stage detector for the sake of accuracy, similar to works in~\cite{ding2019learning,yang2019scrdet}. Unlike the methods in~\cite{zhou2016large,puatruaucean2016jointa,lu2019arc,meng2020arc} which segment or localize the elliptical objects and then fit for ellipse parameters, this framework is an end-to-end learnable pipeline to directly estimate the ellipse five key parameters. Moreover, it is robust to various kinds of degradations in the real fiber detection task compared to traditional ellipse detection algorithms.

\textbf{Learning from synthetic data.} Recently, synthesizing data for CNN training has been used in many computer vision tasks such as crowd counting~\cite{wang2019learning}, semantic segmentation~\cite{zhang2018fully}, person re-identification~\cite{bak2018domain}, etc. However, the bias between synthetic data and real data hurts the generalization ability of the CNNs trained on synthetic data. Domain adaptation is a way of tackling dataset bias by making the synthetic data more realistic-style and minimizing the domain shift at the same time~\cite{zhang2018fully,bak2018domain,wang2019learning,hoffman2017cycada}. In this paper, we generate a synthetic dataset containing large-scale elliptical objects by leveraging some prior knowledge of the elliptical objects. Then we use domain adaptation through CycleGAN~\cite{zhu2017unpaired} to reduce the domain bias between the synthetic and real data.

\textbf{RoI filtering:} Non-Maximum Suppression (NMS) has been an integral part of the object detection. Due to cluttered object proposals generated by Region Proposal Network (RPN) in two-stage detectors, a large number of duplicated bounding boxes decrease the efficiency of CNN training. NMS is generally adopted to solve this challenge by filtering the region proposals according to their classification score. However, it always removes the proposals with lower classification score even if they are with more accurate localization. Recently, soft NMS~\cite{bodla2017soft} and learning NMS~\cite{hosang2017learning} are proposed to optimize the NMS algorithm from the perspective of both the bounding box and the classification score. In this paper, since elliptical object is symmetric, we propose a new RoI ranking algorithm combining classification score and symmetry constraint together and embed them into NMS to refine the localization of region proposals.

\section{Proposed Method \label{sec:Approach}}
The framework of the proposed semi-supervised deep learning method for large-scale elliptical fiber detection is shown in Fig.~\ref{fig:semi}. In this section, we first introduce the semi-supervised deep learning strategy including synthetic data generation, domain adaption and fine-tuning with real data. Then, we explain the detailed components of the proposed elliptical object detector, including the whole structure, the proposed RoI-ellipse learning, and the proposed RoI ranking with the symmetry constraint. In this paper, Regions of Interest (RoIs) have the same meaning as region proposals and we use RoIs for consistency.

\subsection{Semi-supervised deep learning}
\subsubsection{Synthetic data generation}
\begin{figure}[htbh]
\centering
\includegraphics[clip=true, width=0.5\textwidth]{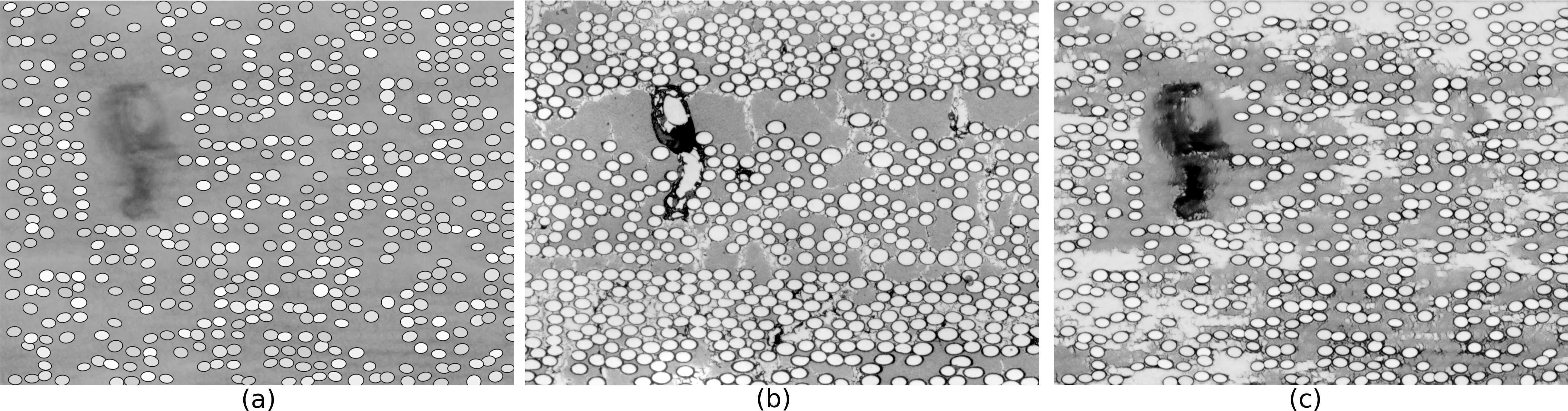} 
\caption{An illustration of the proposed data synthesis: (a) a synthetic ellipse image, (b) a real fiber image, (c) transferred image (a) after the proposed domain adaptation. (a) and (c) share the same ground truth for elliptical fiber detection.}
\label{fig:syn-real}
\end{figure}

Lack of image data with sufficient manual annotations is a common problem in many computer vision tasks. Learning from synthetic data might also achieve good performance compared to the supervised learning strategy~\cite{wang2019learning,zhang2018fully,bak2018domain}. In this work, since it is hard to annotate large-scale elliptical objects in the real dataset, we generate large-scale ellipses by considering prior knowledge. In the fiber dataset from material science in Fig.~\ref{fig:ellipse-vis}, the prior distribution of fibers' major axis, minor axis, angle and region intensity can be learned by fitting a Gaussian distribution for each parameter from a small sampled fiber set (200 annotated real fibers). We then utilize the prior knowledge (fitted Gaussian distributions) to generate thousands of ellipses with different scales. After that, we randomly put them into a synthetic background image which is coarsely obtained by averaging a sequence of real fiber images. Finally, a synthetic dataset of 300 synthetic images (nearly 500 ellipses on each image) is generated, whose ground truth of the five key ellipse parameters is simply recorded during our generation. Figure~\ref{fig:syn-real} (a) and (b) show one example synthetic ellipse image and a real fiber image, respectively. It is clear that the synthetic ellipse image and the real fiber image are quite different in image style, background details and object boundary. Directly training a CNN model on the synthetic data and then testing on the real data might perform poorly due to the domain distribution discrepancy between the synthetic data and the real data. In this paper, the synthetic data is treated as Source domain and the real data is viewed as Target domain. As shown in~\cite{wang2019learning,zhang2018fully,bak2018domain}, the above-mentioned domain distribution discrepancy between Source and Target domains could be relieved by some domain adaptation methods, leading to improved performance.    

\subsubsection{Domain adaptation} 
The domain distribution gap between the synthetic data and real data can be modeled as an image style difference. Recent computer vision researches~\cite{wang2019learning,zhang2018fully,bak2018domain} show that the domain distribution gap can be reduced by translating the image style of synthetic data to the image style of real data. This domain adaptation is implemented in an unsupervised way because we do not have the paired synthetic and real data. For an unpaired image-to-image translation, CycleGAN~\cite{zhu2017unpaired} achieved advanced performance in many computer vision tasks. Here we introduce CycleGAN for the domain adaptation to transfer a synthetic-style image to a real-style image.

The image-to-image transfer in CycleGAN is implemented by training two generators and two adversarial discriminators, as shown in Fig.~\ref{fig:semi}, respectively. We define source domain (synthetic images) as $S$ and target domain (real images) as $T$. With regard to the translation between source domain $S$ and target domain $T$, we define transfer functions ${G}_{1}$ and ${G}_{2}$ as the generators from $S$ to $T$ and from $T$ to $S$, respectively. Meanwhile, two adversarial discriminators $D_{T}$ and $D_{S}$ corresponding to the ${G}_{1}$ and ${G}_{2}$ are defined. Specifically, feeding one image from the domain $S$ to ${G}_{1}$ which acts like a Fully Convolutional Network (FCN) can generate a new image domain $T_{fake}$, and the discriminator $D_{T}$ is to classify whether the new image in domain $T_{fake}$ is a real image from domain $T$ or a fake image generated by that generator. Similarly, ${D}_{S}$ aims to recognize whether an image is a real image from domain $S$ or a fake image generated by ${G}_{2}$. Following ~\cite{zhu2017unpaired}, the total loss function of the domain adaptation is defined as
\begin{equation}\label{eq:gan-total}
\begin{split}
\textit{L}(\textit{G}_{1},\textit{G}_{2},\textit{D}_{S},\textit{D}_{T},\textit{S},\textit{T})
=\textit{L}_{GAN}(\textit{G}_{1},\textit{D}_{T},\textit{S},\textit{T}) \\
+ \textit{L}_{GAN}(\textit{G}_{2},\textit{D}_{S},\textit{T},\textit{S}) 
+ \lambda\textit{L}_{Cyc}(\textit{G}_{1},\textit{G}_{2},S,T), \\
\end{split}
\end{equation}
where $\lambda$ is the weight to balance the adversarial training loss $\textit{L}_{GAN}$ and the cycle consistency loss $\textit{L}_{Cyc}$ in the cycle architecture. $\textit{L}_{Cyc}$ is to keep the transfers from $S$ to $T$ and from $T$ to $S$ cycle-consistent, which is defined as
\begin{equation}
\begin{split}
\textit{L}_{cyc}(\textit{G}_{1},\textit{G}_{2},S,T)=\textit{E}_{\textit{i}_{S}\sim\textit{I}_{S}}[||\textit{G}_{2}(\textit{G}_{1}(\textit{i}_{S}))-\textit{i}_{S}||_{1}] \\
+\textit{E}_{\textit{i}_{T}\sim\textit{I}_{T}}[||\textit{G}_{1}(\textit{G}_{2}(\textit{i}_{T}))-\textit{i}_{T}||_{1}],
\end{split}
\end{equation}
where $\textit{i}_{S}$ $\in$ $\textit{I}_{S}$ and $\textit{i}_{T}$ $\in$ $\textit{I}_{T}$ represent any images in $S$ and $T$, respectively. The adversarial training loss function is defined as
\begin{equation}\label{eq:gan_loss}
\begin{split}
    \textit{L}_{GAN}(\textit{G}_{1},\textit{D}_{T},\textit{S},\textit{T}) =\textit{E}_{\textit{i}_{T}\sim\textit{I}_{T}}[\log(\textit{D}_{T}(\textit{i}_{T}))]\\
    +\textit{E}_{\textit{i}_{S}\sim\textit{I}_{S}}[\log(1-\textit{D}_{T}(\textit{G}_{1}(\textit{i}_{S}))]. 
\end{split}
\end{equation}
The training of these generators and discriminators aims to solve the optimization problem of  
\begin{equation}\label{eq:gan_opti}
    \textit{G}_{1}^*, \textit{G}_{2}^* =\arg 
    \operatorname*{min}_{G_{1},G_{2}}\operatorname*{max}_{D_{S},D_{T}} \textit{L}(\textit{G}_{1},\textit{G}_{2},\textit{D}_{S},\textit{D}_{T},\textit{S},\textit{T}). 
\end{equation}

After solving Eq.~(\ref{eq:gan_opti}) by gradient descent and back propagation, the learned generator $\textit{G}_{1}^*$ can be used to transfer the synthetic-style image to the corresponding real-style image. Figure~\ref{fig:syn-real}(c) shows the transferred (fake) image from the synthetic image after domain adaptation. We can see that the image style, background details and object boundary of the transferred image are much more similar to the real image than the synthetic image. Note that the transferred image shares the same ground truth with the original synthetic image for elliptical fiber detection. This way, the domain distribution discrepancy between the synthetic data and the real data is reduced and the transferred images can be used to train a more robust elliptical fiber detector.

\subsubsection{Fine-tuning based generalization} 
To improve the generalization ability of the trained CNNs, we also fine-tune the network trained after domain adaptation on a small number of manually annotated real data. 

\subsection{Elliptical fiber detector}
In this paper, the elliptical fiber detector is developed from the two-stage object detector Faster R-CNN~\cite{ren2015faster} with VGG16~\cite{simonyan2014very} as backbone, as shown in Fig.~\ref{fig:semi}. The Faster R-CNN model has two phases: RPN~\cite{ren2015faster} phase and R-CNN (Region-CNN) phase. The RPN phase is mainly learning to generate RoIs for objects, and the R-CNN phase is to perform the object classification and refine the localization of RoIs.      

\subsubsection{RoI-ellipse learning}
\begin{figure}[htbh]
\centering
\includegraphics[width=1.0\columnwidth]{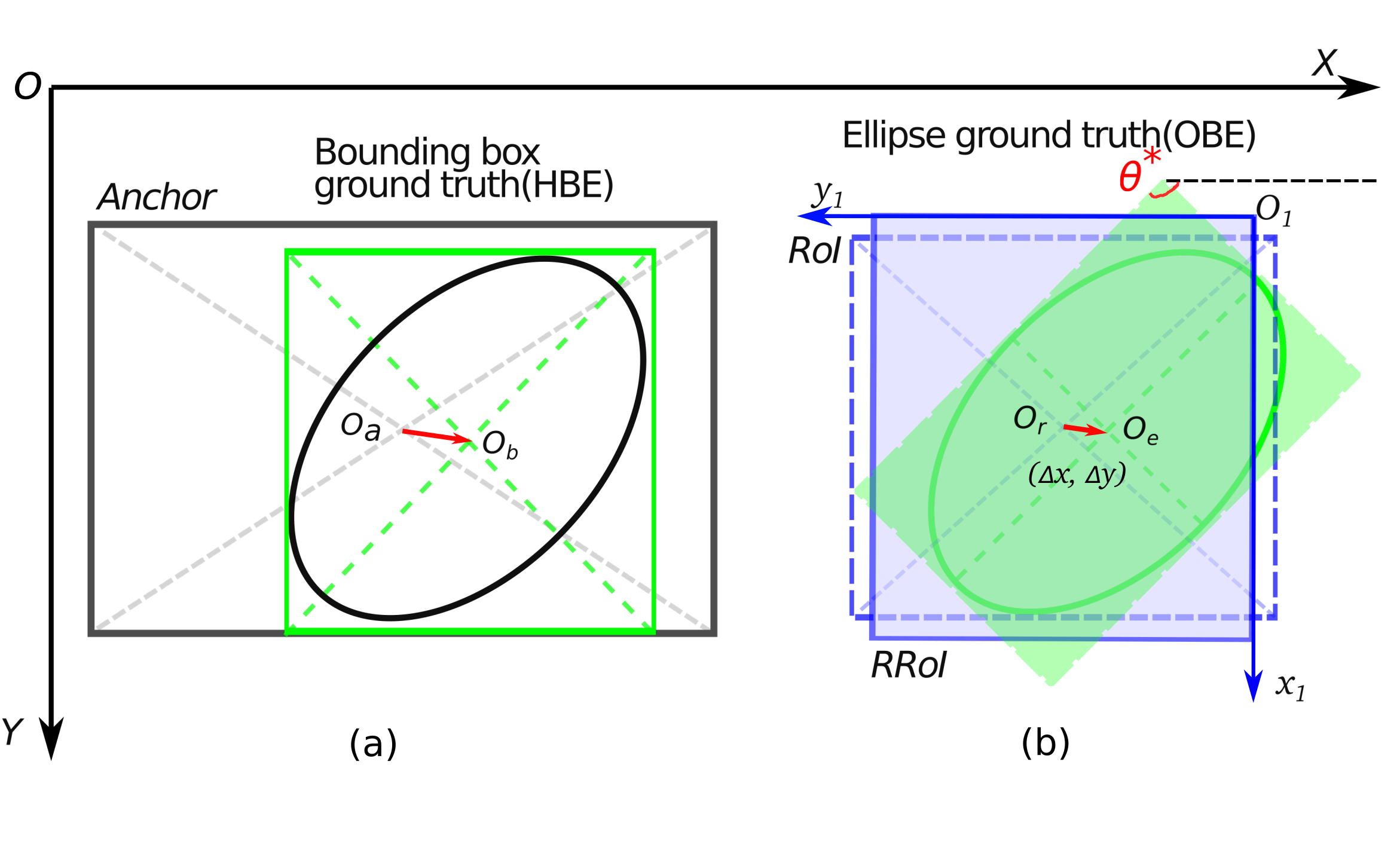} 
\caption{Regression processes of RoI-ellipse learning: (a) regression from anchor to HBE in the RPN phase, (b) regression from RRoI to OBE in the R-CNN phase. $O_{a}$, $O_{b}$, $O_{r}$ and $O_{e}$ are the center of anchor, HBE, RRoI and OBE, respectively. ($\triangle x$, $\triangle y$) is the center offset.}
\label{fig:RoI-ellipse}
\end{figure}

Inspired by \cite{ding2019learning}, the RoI-ellipse learning is implemented in the form of predicting the five key ellipse parameters $[\textit{x}^*, \textit{y}^*, \textit{2R}^*, \textit{2r}^*, \textit{$\theta$}^*]$ for elliptical fibers, which can also be viewed as the same parameters of the $\textit{oriented bounding box over ellipse}$ (OBE), as shown by the green box in Fig.~\ref{fig:RoI-ellipse}(b). Specifically, OBE tightly bounds the ellipse with four sides parallel to the major/minor axes of the ellipse and the orientation $\textit{$\theta$}^*$ is the angle between the major axis and the horizontal axis. In this section, we include two regression processes for the proposed RoI-ellipse learning in RPN phase and R-CNN phase, respectively.

After multiple anchors are generated, the goal of the regression in the RPN phase is to initially learn the $\textit{horizontal bounding box over ellipse}$ (HBE), as shown by the green box in Fig.~\ref{fig:RoI-ellipse}(a), rather than directly learn the ellipse five parameters from feature map of an anchor. HBE tightly bounds the ellipse with four sides parallel to image sides. The regression offsets from an anchor to the target HBE are the same as that in the standard Faster R-CNN~\cite{ren2015faster}. Note that when matching an anchor with all the HBEs to recognize whether it contains an ellipse or not, we need to transform the ellipse to its HBE by mapping the five key ellipse parameters to four HBE parameters $[\textit{x}, \textit{y}, \textit{w}, \textit{h}]$, where $(x,y)$ is the HBE center and $w$ and $h$ are its width and height. Through CNN training, RPN outputs a list of RoIs, each of which contains an elliptical object even though each RoI may not capture the underlying ellipse accurately: RoI probably doesn't fit the elliptical object very well as shown in Fig.~\ref{fig:RoI-ellipse}(b).

In general, the goal of the regression in R-CNN stage is to refine the localization of each RoI to fit the underlying elliptical objects better. In this work, however, the goal of regression in this phase is to detect the five key parameters of the ellipse, which can be achieved by learning the corresponding OBE from the feature map of an $\textit{rotated RoI}$ (RRoI). Inspired by~\cite{ding2019learning,xu2019geometry}, we rotate each RoI by 90 degrees in a clockwise direction to obtain its RRoI in order to reduce the geometry variation of rotation, as shown by the blue box in Fig.~\ref{fig:RoI-ellipse}(b). Following~\cite{ding2019learning}, we use a few fully connected layers for inferring the five parameters of ellipse given the linear transform between an RRoI and its corresponding OBE. After projecting center offset ($\triangle x$, $\triangle y$) from the global coordinate system $\textit{XOY}$ into the local coordinate system $\textit{x}_{1}\textit{O}_{1}\textit{y}_{1}$, as shown in Fig.~\ref{fig:RoI-ellipse}(b), we define the regression offsets from an RRoI to its OBE in this work as
\begin{equation}
t_x^{*}=\frac{1}{w^r}(y^{*}-y^r)
\end{equation}
\begin{equation}\label{eq:reg-error}
t_y^{*}=\frac{1}{h^r}[-(x^{*}-x^r)]
\end{equation}
\begin{equation}
t_w^{*}=\log\frac{w^*}{w^r}
\end{equation}
\begin{equation}
t_h^{*}=\log\frac{h^*}{h^r}
\end{equation}
\begin{equation}
t_\theta^{*}=\frac{1}{\pi/2}(\theta^*-\frac{\pi}{2}), 
\end{equation}
where [$x^r$, $y^r$, $w^r$, $h^r$] are 2D center location, width and height of the RRoI and  [$x^{*}$, $y^{*}$, $w^{*}$, $h^{*}$, $\theta^{*}$] are the 2D center location, major-axis length, minor-axis length and orientation angle of the ground-truth ellipse.

Different from~\cite{ding2019learning}, the value of target angle $\theta^{*}$ ranges from 0 to $\pi$ due to the symmetry of ellipse, we set the the initial angle of RRoI to $\frac{\pi}{2}$ in order to reduce the geometry variation of rotation. Because large geometry variation could harm the detection performance~\cite{xu2019geometry}. The target offset $t_\theta^{*}$ is normalized to [-1, 1]. We only use one-step R-CNN regression, instead of two-step R-CNN regression proposed in~\cite{ding2019learning}, to detect ellipse key parameters for efficiency. During the training, the Smooth $L_1$ loss function~\cite{girshick2014rich} is adopted as the regression loss. 

\subsubsection{RoI ranking with symmetry constraint}
NMS is an indispensable step of the R-CNN-based object detection~\cite{ren2015faster,he2017mask}, where category independent RoIs are generated by RPN and then fed to two sub-networks: a classification sub-network and a regression sub-network. The former outputs a category-specific classification score and the latter
performs regression to refine the localization of RoIs to the real localization of the corresponding objects. In this process, NMS acts like a filter to keep only high-quality RoIs for the subsequent sub-networks. 

Traditionally, an RoI with high classification score is assumed to be of high quality. A typical NMS process consists of the following steps: 1) Rank all RoIs based on their classification score $S_{obj}$, from the highest down to the lowest. 2) Select the RoI with the highest score $RoI_{max}$ and remove all the other RoIs with Intersection over Union (IoU) greater than a threshold with $RoI_{max}$. 3) Append the selected RoI to the final RoIs set. 4) In the remaining RoIs, repeat the first three steps until no RoIs remain. However, in practice, the classification score may not well reflect the quality of RoIs, i.e., RoIs that are with accurate localization may show lower score and not be selected in NMS.

\begin{figure}[htbh]
\centering
\includegraphics[width=1.0\columnwidth]{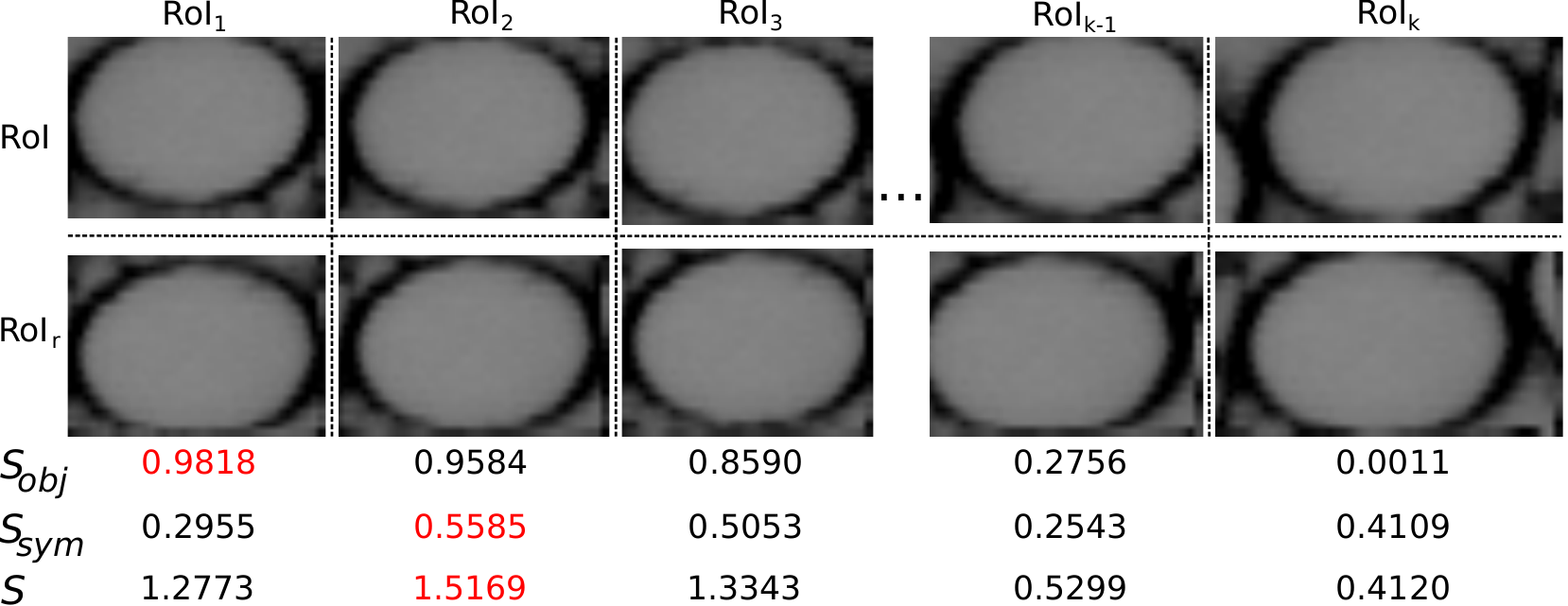}
\caption{Sample results of the proposed RoI ranking. 
Top: RoIs with $>0.7$ IoU to the selected $\text{RoI}_{max}$. Bottom: 180-degree rotation
of the RoIs in the top row. Classification score, SSIM-based symmetry score and the combined score are shown below each RoI.}
\label{fig:nmsi}
\end{figure}

To alleviate this problem, we propose a new RoI ranking operation by considering the object symmetry. We know that the ideal ellipse shape is symmetric when rotating around the center by 180 degrees. We can evaluate such symmetry for each RoI and use it to help the RoI selection. Specifically, for each RoI, we rotate it by 180 degrees around its center and then compute the Structural Similarity Index Measure (SSIM) ~\cite{wang2004image} between the original RoI and its rotated version. SSIM-based symmetry score is denoted as $S_{sym}\in [0,1]$: the larger the $S_{sym}$ 
is, the more symmetric the RoI can be. We can then define a combined score
\begin{equation}
S = S_{obj} + \lambda S_{sym}, 
\end{equation}
for RoI selection in NMS. Here $\lambda$ is a balance factor, which we set to 1 in our experiments. Other steps of NMS remain the same. As shown in Fig.~\ref{fig:nmsi}, using standard NMS, $\text{RoI}_{1}$ has the highest classification score: it is selected in NMS and then $\text{RoI}_{j}(j=2,3,...,k)$ are removed. However, using the proposed RoI ranking, $\text{RoI}_{2}$ will be selected since it has the highest combined score $S$. 
In this case, $\text{RoI}_{2}$ fits the underlying object better than $\text{RoI}_{1}$ as shown in Fig.~\ref{fig:nmsi}. 

\section{Experiments \label{sec:Experiments} }
\subsection{Dataset}
In this research, we generate synthetic data for training and also fine-tune and evaluate our model on real data.  For the synthetic dataset, 300 images with resolution $1,292\times968$ are generated and each image is embedded with about 500 ellipses. For the real dataset, we use the public FRC image  dataset~\cite{yu2016groupwise,zhou2016large} obtained by cross-sectioning the S200 material. In this real public dataset, there are three subsets and each subset has 100 microscopic images/slices. Each image's resolution is $1,292\times968$ with about 600 elliptical objects (fibers). Image slices present blur and stain degradations caused by the manufacturing process. The fibers on each image slice are arranged in a large-scale layout.
The FRC image dataset does not have ground truth annotations for each fiber's five ellipse parameters, therefore we  carefully annotated 36 images for each fiber's five ellipse parameters.  Note that it takes 1 to 2 hours to annotate one image with about 600 elliptical objects (fibers). In this 36 carefully annotated images, 9 real images are used for fine-tuning the model trained on synthetic images after domain adaptation and the remaining 27 real images are evenly divided into three testing sets $\textit{Data}_{1}$, $\textit{Data}_{2}$ and $\textit{Data}_{3}$. There are 4, 1, 3 and 5 images with degradations out of every 9 real images in the fine-tune training set, $\textit{Data}_{1}$, $\textit{Data}_{2}$ and  $\textit{Data}_{3}$, respectively. We also construct two subsets ${Det}_{c}$ and ${Det}_{d}$ from the public FRC image dataset with 300 real images for bounding-box-level comparison. ${Det}_{d}$ is a subset of 40 images with degradation effects like stains or blurs. The remaining images of the public FRC image dataset, 260 clean images without any degradation effects, constitute the subset ${Det}_{c}$. 

\subsection{Experimental setting}
We use VGG16 as our backbone for feature extraction and the proposed method is implemented in PyTorch. We utilize Stochastic Gradient Descent (SGD) with a weight decay of 0.1 and momentum of 0.9 to optimize all models and set the initial learning rate to 0.001, reduce it by the weight decay after 20 epochs. All the experiments are implemented on 1 NVIDIA GeForce GTX 1080Ti GPU.

For synthetic data, there are 150,000 ($300\times500$) target objects for one epoch training, we train 40 epochs with batch size of 4 
either before or after domain adaptation, which does not have over-fitting in our experiments. When fine-tuning the model with 9 real images, there are 5,400 ($9\times600$) target objects for one epoch training, we run 60 epochs. During the domain adaptation between the synthetic data and the real data, since GANs are typically hard to train, we reduce the discriminators of CycleGAN to 1 convolution layer in order to avoid the shift of positions of the generated ellipses after domain adaptation in this fiber detection task. The other parameters  follow the default setting in CycleGAN~\cite{zhu2017unpaired}.
After obtaining the detection result, NMS is also utilized here to filter duplicated ellipses regressed to the same object. Instead of using the region-based IoU here, we select the detected ellipse with the maximum combined score for each object and then remove any other detected ellipses with distance less than 20 pixels to the selected one.      

For performance evaluation of elliptical fiber detection, direct comparing the values of the predicted five parameters against their ground truth is not desirable since there are five parameters and it is nontrivial to combine them into a single comprehensive evaluation metric.

In this paper, we follow previous works~\cite{fornaciari2014fast,jia2017fast} by adopting pixel-based IoU as the criterion to evaluate the performance of elliptical fiber detection. We employ two ways to evaluate the proposed method: 1) ellipse-level evaluation and 2) bounding-box-level evaluation, both in the format of popular detection metrics (Precision, Recall and F-measure). For the ellipse-level evaluation, we quantify how many ellipses are detected by measuring the pixel-based IoU of detected ellipses and ground-truth ellipses. If the pixel-based IoU of a predicted ellipse with its corresponding ground truth ellipse is more than 0.5, we regard this detection as a true positive. For the bounding-box-level evaluation, we transform the detected ellipse to its tightest horizontal bounding box (HBE) first and then compute those detection metrics based on region-based IoU.

\subsection{Ellipse level evaluation}

\begin{table}[htbp]
\begin{centering}
\caption{Experimental results of elliptical fiber detection in terms of ellipse-level evaluation metrics on $Data_1$,  $Data_2$ and $Data_3$ of the real public FRC image dataset~\cite{yu2016groupwise,zhou2016large}.  There are 1, 3 and 5 degraded images out of 9 images on $Data_1$,  $Data_2$ and $Data_3$, respectively. F: F-measure, P: Precision, R: Recall.}
\footnotesize
\begin{tabular}{c|ccc|c}
\hline 
\hline
\cline{0-4}  $Data_1$ & $\text{F(\%)}$ & $\text{P(\%)}$ & $\text{R(\%)}$ & $\text{ML1}_{deg}(\degree)$ \tabularnewline
\cline{0-4} 
$\text{ELSD~\cite{puatruaucean2016jointa}}$ & 73.59 & 52.94 & 83.35 & \textit{2.73}\tabularnewline
$\text{Lu~\cite{lu2019arc}}$ & 73.92 & 94.46 &  69.40 & \textbf{2.46} \tabularnewline
$\text{AAMED~\cite{meng2020arc}}$ & 94.24 & \textit{94.65} &  94.11 & 3.09 \tabularnewline
$\text{EMMPMH~\cite{zhou2016large}}$ & \textit{96.83} &  93.83 & \textit{97.77} & 6.86 \tabularnewline
$\text{Faster R-CNN~\cite{ren2015faster}}$ & 93.56 &  {94.39} & 93.31 & 16.18 \tabularnewline
$\text{Proposed}$ & {\textbf{97.91}} & {\textbf{95.08}} & {\textbf{98.80}} & {5.19} \tabularnewline
\hline
\hline
\cline{0-4} $Data_2$ & $\text{F(\%)}$ & $\text{P(\%)}$ & $\text{R(\%)}$ & $\text{ML1}_{deg}(\degree)$ \tabularnewline
\cline{0-4} 
$\text{ELSD~\cite{puatruaucean2016jointa}}$ & 73.23 & 52.43 & 83.12 & \textit{2.51} \tabularnewline
$\text{Lu~\cite{lu2019arc}}$ & 72.65 & \textit{94.33} &  67.96 & \textbf{2.26} \tabularnewline
$\text{AAMED~\cite{meng2020arc}}$ & 93.37 & 94.30 &  93.09 & 2.87 \tabularnewline
$\text{EMMPMH~\cite{zhou2016large}}$ & \textit{94.67} &  93.21 & 95.12 & 6.46 \tabularnewline
$\text{Faster R-CNN~\cite{ren2015faster}}$ & 92.93 &  93.89 & \textit{92.64} & 15.58 \tabularnewline
$\text{Proposed}$ & {\textbf{97.74}} & {\textbf{94.68}} & {\textbf{98.70}} & {4.93} \tabularnewline
\hline
\hline
\cline{0-4} $Data_3$ & $\text{F(\%)}$ & $\text{P(\%)}$ & $\text{R(\%)}$ & $\text{ML1}_{deg}(\degree)$ \tabularnewline
\cline{0-4} 
$\text{ELSD~\cite{puatruaucean2016jointa}}$ & 73.45 & 53.43 & 82.76 & \textit{3.51} \tabularnewline
$\text{Lu~\cite{lu2019arc}}$ & 71.01 & \textit{95.62} &  65.91 & \textbf{3.08} \tabularnewline
$\text{AAMED~\cite{meng2020arc}}$ & 92.41 & 95.53 &  91.52 & 4.05 \tabularnewline
$\text{EMMPMH~\cite{zhou2016large}}$ & {91.68} &  {93.98} & {91.01} & 7.12 \tabularnewline
$\text{Faster R-CNN~\cite{ren2015faster}}$ & \textit{93.11} &  {95.15} & \textit{92.52} & 14.22 \tabularnewline
$\text{Proposed}$ & {\textbf{98.05}} & {\textbf{95.67}} & {\textbf{98.79}} & {4.29} \tabularnewline
\hline
\hline
\end{tabular} \label{tab:eval_seg}
\par\end{centering}
\end{table}

First, we compare the performance of the proposed method with fiber detection method EMMPMH~\cite{zhou2016large}, ellipse detection methods: ELSD~\cite{puatruaucean2016jointa}, Lu~\cite{lu2019arc} and AAMED~\cite{meng2020arc} and deep learning based method Faster R-CNN~\cite{ren2015faster}. Utilizing Faster R-CNN as a baseline method for elliptical object detection, we regress the ellipse parameters $[x,y,2R,2r]$ by keeping consistent with the target offsets in 
the work~\cite{ren2015faster}. For the orientation angle detection, we regress the target angle directly from an RoI to an ellipse. We train on fine-tune training dataset (9 real images) directly without the help of synthetic data from the scratch to avoid over-fitting. The result is shown in Table~\ref{tab:eval_seg}. 

\begin{figure*}[htbh]
\centering
\includegraphics[width=0.9\textwidth]{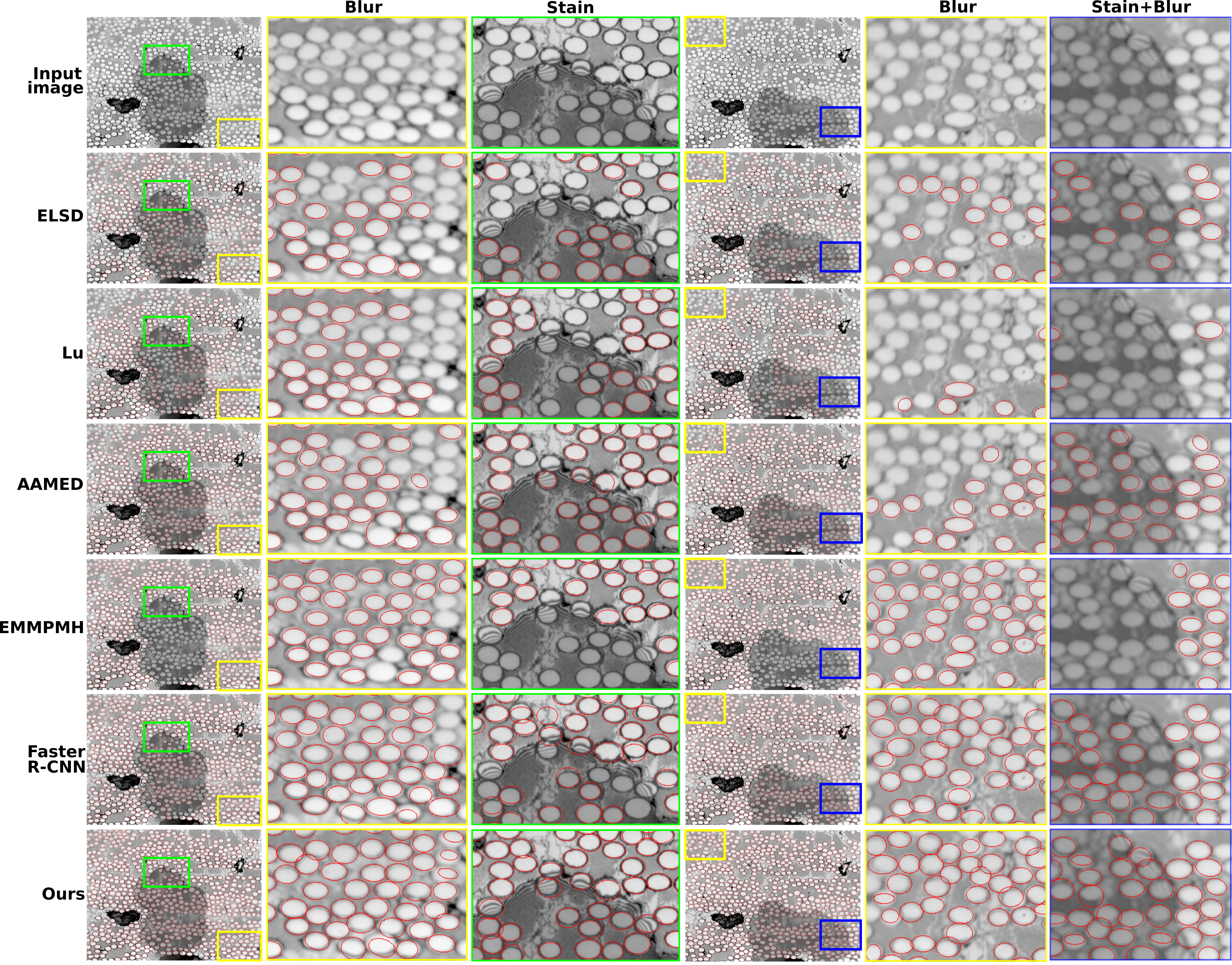} 
\caption{Visualization results of the elliptical fiber detection: original real images, detection results by ELSD~\cite{puatruaucean2016jointa}, Lu~\cite{lu2019arc}, AAMED~\cite{meng2020arc}, EMMPMH~\cite{zhou2016large}, Faster R-CNN~\cite{ren2015faster} and the proposed method. The bounding box in the image is amplified in the next two columns for better visualized comparisons. Note that many objects are missed by using ELSD, Lu, AAMED and EMMPMH, especially in the degraded regions. }
\label{fig:elsd-emmpmh}
\end{figure*}

It shows that the proposed method performs better than the comparison methods in terms of Precision, Recall and the overall F-measure. 
Specifically, on $Data_3$, the proposed method outperforms the state-of-the-art ellipse detection method AAMED by over 5\% in F-measure and 7\% in Recall. ELSD doesn't perform well in the form of Precision, since it mainly fits and validates the grouped line segments locally without considering the global information, leading to more false positives. Lu enhanced ELSD by robust arc-support line segments grouping and effective ellipse setting locally and globally. It improves Precision more than 40\% while reducing Recall largely compared to ELSD, resulting in less than 75\% F-measure similar to ELSD. The reason is that Lu's approach is sensitive to image degradations: blurs and stains, as shown in Fig.~\ref{fig:elsd-emmpmh}. AAMED performs better than ELSD and Lu's approach with over 92\% F-measure, since it has more effective representation via digraph-based AAM and acquires more true ellipses via the bidirectional search strategy. Fiber detection method EMMPMH fails to detect more true fibers due to sensitivity to image degradations. 
The proposed method performs better compared to baseline method Faster R-CNN with the help of synthetic data, domain adaptation, effective RoI-ellipse learning and RoI ranking with symmetry constraint, which improve the performance by about 5\% in F-measure and over 6\% in Recall.

Moreover, the more degraded images existing in testing set, the better the deep learning based methods perform than the other detectors. For example, the proposed method outperforms the state-of-the-art fiber detection method EMMPMH by over 6\% increase in F-measure for $Data_3$ compared to 1\% for $Data_1$. In particular, it improves the Recall significantly by more than 7.7\% on $Data_3$ compared to 1.0\% on $Data_1$, since CNN training can supervisedly learn deep features to handle image degradations than the handcrafted-feature based EMMPMH.

We also evaluate the performance of orientation detection by measuring the difference of the orientation angle of a predicted ellipse and its corresponding ground-truth ellipse. We only evaluate the true-positive detected ellipses, which is consistent for all the comparison methods. The evaluation metric is Mean L1 distance in the format of degree ($\text{ML1}_{deg}$), shown in Table~\ref{tab:eval_seg}. 
We can see that Lu's approach performs best for orientation prediction but with lowest Recall, while the proposed method performs better for Recall and achieves relatively satisfied orientation prediction.

Visualization results are shown in Fig.~\ref{fig:elsd-emmpmh}. We can see that the Lu detector doesn't perform well in various kinds of image degradations compared to other detection algorithms. ELSD and AAMED performs well in regions with stain degradation, while poorly when the image becomes blurred. AAMED detects more true positive ellipses than ELSD. EMMPMH detector mainly fails in regions with stains instead of blurs, while the proposed method performs better than other methods in regions with various kinds of image degradations due to CNN's powerful feature learning ability. Faster R-CNN detector misses some fibers due to training with limited data. 
\begin{table}[htbp]
\begin{centering}
\caption{Experimental results of elliptical fiber detection in terms of bounding-box level evaluation metrics on $Det_c$ (260 images) and $Det_d$ (40 images) of the real public FRC image dataset~\cite{yu2016groupwise,zhou2016large}.}
\footnotesize
\begin{tabular}{c|ccc}
\hline 
\hline
\cline{0-3} ${Det}_{c}$  & $\text{F-measure(\%)}$ & $\text{Precision(\%)}$ & $\text{Recall(\%)}$ \tabularnewline
\cline{0-3} 
$\text{ELSD~\cite{puatruaucean2016jointa}}$ & 52.46 &  43.35 & 66.42 \tabularnewline
$\text{Lu~\cite{lu2019arc}}$ & 82.65 &  99.22 & 70.82 \tabularnewline
$\text{AAMED~\cite{meng2020arc}}$ & 95.61 &  \textbf{99.60} & 91.92 \tabularnewline
$\text{EMMPMH~\cite{zhou2016large}}$ & \textit{97.00} &  98.85 & \textbf{95.22} \tabularnewline
$\text{Faster R-CNN~\cite{ren2015faster}}$ & 92.54 &  95.33 & 89.90 \tabularnewline
$\text{Proposed}$ & \textbf{97.19} & \textit{99.23} & \textbf{95.22}  \tabularnewline
\hline
\hline
\cline{0-3} ${Det}_{d}$ & $\text{F-measure(\%)}$ & $\text{Precision(\%)}$ & $\text{Recall(\%)}$\tabularnewline
\cline{0-3} 
$\text{ELSD~\cite{puatruaucean2016jointa}}$ & 61.15 &  50.92 & 76.54  \tabularnewline
$\text{Lu~\cite{lu2019arc}}$ & 77.04 &  97.50 & 63.68 \tabularnewline
$\text{AAMED~\cite{meng2020arc}}$ & 93.47 &  \textit{98.90} & 88.59 \tabularnewline
$\text{EMMPMH~\cite{zhou2016large}}$ & \textit{94.45} &  97.56 & \textit{91.53}  \tabularnewline
$\text{Faster R-CNN~\cite{ren2015faster}}$ & 91.83 &  96.08 & 87.94  \tabularnewline
$\text{Proposed}$ & \textbf{96.99} & \textbf{99.48} & \textbf{94.62}  \tabularnewline
\hline
\hline
\end{tabular} \label{tab:eval_bbx}
\par\end{centering}
\end{table}

\subsection{Bounding-box level evaluation}
Most of current object detection methods are evaluated by comparing the region-based IoU of HBEs and ground-truth bounding boxes. In this work, we also evaluate our elliptical fiber detection result with other detection methods by comparing the bounding boxes. The result is reported in Table~\ref{tab:eval_bbx}.  We can see that all the detection performance decreases on ${Det}_{d}$ dataset, except for ELSD, especially in Recall. The proposed method achieves the best performance on both ${Det}_{c}$  and ${Det}_{d}$ subsets. For detection on clean images, our method obtains comparable performance with EMMPMH, both of which achieves F-measure about 97\% and Recall 95.22\%. For detection on images with degradation effects, the proposed method outperforms the second best method EMMPMH with about 2.5\% increase of F-measure, about 2\% increase of Precision, and about 3\% increase of Recall. The baseline deep learning based method performs poorly in bounding box level evaluation, since it is trained with limited data and its orientation angle detection is not accurate enough. 

\subsection{Ablation Study}
In this section, we evaluate the contribution of each step in the proposed method: 1) training on the synthetic images, denoted as $\text{Proposed}_{Syn}$; 2) training on the synthetic images after the CycleGAN-based domain adaptation, denoted as $\text{Proposed}_{Syn+D}$; 3) fine-tuning the $\text{Proposed}_{Syn+D}$ on the real images, denoted as $\text{Proposed}_{Syn+D+F}$; 4) fine-tuning the $\text{Proposed}_{Syn+D}$ on the real images with the RoI ranking with the symmetry constraint, denoted as $\text{Proposed}_{Syn+D+F+R}$. For the standard semi-supervised baseline, we fine-tune the $\text{Proposed}_{Syn}$ on the real images directly, denoted as  $\text{Proposed}_{Syn+F}$. The methods $\text{Proposed}_{Syn}$ and  $\text{Proposed}_{Syn+F}$ are considered as the baseline algorithms in this experimental setting. 

\begin{table}[htbp]
\begin{centering}
\caption{Ablation study on the use of synthetic data, domain adaptation, fine-tuning, and RoI ranking with the symmetry constraint on $Data_1$, $Data_2$ and $Data_3$ of the real public FRC image dataset~\cite{yu2016groupwise,zhou2016large} in terms of ellipse-level evaluation metrics.}
\footnotesize
\begin{tabular}{c|ccc}
\hline 
\hline
\cline{0-3} $Data_1$ & $\text{F-measure(\%)}$ & $\text{Precision(\%)}$ & $\text{Recall(\%)}$ \tabularnewline
\cline{0-3} 
 $\text{Proposed}_{Syn}$ &76.21 & 92.57 & 72.37  \tabularnewline
 $\text{Proposed}_{Syn+F}$ & 96.15 & \textit{96.17} & 96.14 \tabularnewline
 \cline{0-3} 
 $\text{Proposed}_{Syn+D}$ &86.09 & \textbf{96.67} & 83.35 \tabularnewline
 $\text{Proposed}_{Syn+D+F}$ & \textit{97.77} & 95.96 & \textit{98.33}  \tabularnewline
 $\text{Proposed}_{Syn+D+F+R}$ & \textbf{97.91} & 95.08 & \textbf{98.80} \tabularnewline
\hline
\hline
\cline{0-3} $Data_2$ & $\text{F-measure(\%)}$ & $\text{Precision(\%)}$ & $\text{Recall(\%)}$\tabularnewline
\cline{0-3} 
 $\text{Proposed}_{Syn}$ &77.08 & 92.18 & 73.47  \tabularnewline
 $\text{Proposed}_{Syn+F}$ & 95.72 & \textit{95.78} & 95.71 \tabularnewline
 \cline{0-3} 
 $\text{Proposed}_{Syn+D}$&85.25 & \textbf{95.97} & 82.48 \tabularnewline
 $\text{Proposed}_{Syn+D+F}$ & \textit{97.30} & 95.29 & \textit{97.93}  \tabularnewline
 $\text{Proposed}_{Syn+D+F+R}$ & \textbf{97.74} & 94.68 & \textbf{98.70} \tabularnewline
\hline
\hline
\cline{0-3} $Data_3$ & $\text{F-measure(\%)}$ & $\text{Precision(\%)}$ & $\text{Recall(\%)}$\tabularnewline
\cline{0-3} 
 $\text{Proposed}_{Syn}$&69.92 & 93.42 & 65.01 \tabularnewline
 $\text{Proposed}_{Syn+F}$ & 95.86 & \textit{97.10} & 95.49 \tabularnewline
 \cline{0-3} 
 $\text{Proposed}_{Syn+D}$&78.78 & \textbf{97.82} & 74.43 \tabularnewline
 $\text{Proposed}_{Syn+D+F}$ & \textit{97.30} & 96.56 & \textit{97.53}\tabularnewline
 $\text{Proposed}_{Syn+D+F+R}$ & \textbf{98.05} & 95.67 & \textbf{98.79} \tabularnewline
\hline
\hline
\end{tabular} \label{tab:ablation_study}
\par\end{centering}
\end{table}

The corresponding result of each step of the proposed method on $Data_1$, $Data_2$ and $Data_3$ in terms of ellipse level evaluation metrics is reported in Table~\ref{tab:ablation_study}. We can clearly see the positive effect of each step with the increasing overall F-measure performance. Taking the subset $Data_3$ as an example, using only synthetic data obtains 69.92\% F-measure, and adding domain adaptation increases it to 78.78\% F-measure, and continually adding the fine-tuning step increases it to 97.30\% F-measure, and continually adding the RoI ranking improves it to 98.05\% F-measure. Same increase trend happens in the subsets $Data_1$ and $Data_2$. Comparing to the baseline algorithms $\text{Proposed}_{Syn}$ and  $\text{Proposed}_{Syn+F}$, the proposed method could improve the standard semi-supervised learning method by involving the steps of domain adaptation and RoI ranking. It turns out that fine-tuning on real images directly is not good enough to cover the performance gain provided by the proposed method.

\begin{table}[htbp]
\begin{centering}
\caption{Ablation study on the use of synthetic data, domain adaptation, fine-tuning, and RoI ranking with the symmetry constraint on $Data_1$, $Data_2$ and $Data_3$ of the real public FRC image dataset~\cite{yu2016groupwise,zhou2016large} in terms of orientation evaluation metrics.}
\footnotesize
\begin{tabular}{c|cccc}
\hline 
\hline
\cline{0-3} $Data_1$ & $\text{ML1}_{rad}$ & $\text{MSE}_{rad}$ & $\text{ML1}_{deg}(\degree)$\tabularnewline
 \cline{0-3} 
 $\text{Proposed}_{Syn}$ & 0.308 & 0.228 & 17.641
 \tabularnewline
 $\text{Proposed}_{Syn+F}$ & 0.116 & 0.021 & 6.639\tabularnewline
  \cline{0-3} 
 $\text{Proposed}_{Syn+D}$ & 0.197 & 0.118 & 11.279\tabularnewline
 $\text{Proposed}_{Syn+D+F}$ & \textit{0.096} & \textit{0.016} & \textit{5.527} \tabularnewline
 $\text{Proposed}_{Syn+D+F+R}$ & \textbf{0.091} & \textbf{0.014} & \textbf{5.195} \tabularnewline
\hline
\hline
\cline{0-3} $Data_2$ & $\text{ML1}_{rad}$ & $\text{MSE}_{rad}$ & $\text{ML1}_{deg}(\degree)$\tabularnewline

 \cline{0-3} 
 $\text{Proposed}_{Syn}$& 0.293 & 0.215 & 16.759\tabularnewline
 $\text{Proposed}_{Syn+F}$ & 0.114 & 0.019 & 6.551\tabularnewline
  \cline{0-3} 
 $\text{Proposed}_{Syn+D}$ & 0.204 & 0.124 & 11.702\tabularnewline
 $\text{Proposed}_{Syn+D+F}$ & \textit{0.092} & \textit{0.014} & \textit{5.291} \tabularnewline
 $\text{Proposed}_{Syn+D+F+R}$ & \textbf{0.086} & \textbf{0.012} & \textbf{4.933}\tabularnewline
\hline
\hline
\cline{0-3} $Data_3$ & $\text{ML1}_{rad}$ & $\text{MSE}_{rad}$ & $\text{ML1}_{deg}(\degree)$\tabularnewline

 \cline{0-3} 
 $\text{Proposed}_{Syn}$& 0.329 & 0.245 & 18.823 \tabularnewline
 $\text{Proposed}_{Syn+F}$ & 0.088 & 0.014 & 5.027 \tabularnewline
  \cline{0-3} 
 $\text{Proposed}_{Syn+D}$& 0.201 & 0.119 & 11.495 \tabularnewline
 $\text{Proposed}_{Syn+D+F}$ & \textit{0.079} & \textit{0.012} & \textit{4.553} \tabularnewline
 $\text{Proposed}_{Syn+D+F+R}$ & \textbf{0.075} & \textbf{0.011} & \textbf{4.293} \tabularnewline
\hline
\hline
\end{tabular} \label{tab:ablation_study_ori}
\par\end{centering}
\end{table}

We also report the contribution of each step of the proposed method on $Data_1$, $Data_2$ and $Data_3$ in terms of ellipse orientation evaluation metrics: Mean L1 distance in the format of radian, Mean Squared Error in the format of radian and Mean L1 distance in the format of degree, shown in Table~\ref{tab:ablation_study_ori}. It also verifies the effectiveness of the proposed method.



\section{Conclusions and Discussion\label{sec:Conclusions}}
In this paper, we proposed a semi-supervised detection method that combines the synthetic data generation, domain adaptation, fine-tuning, and RoI ranking with the symmetry constraint for large-scale elliptical fiber detection. It significantly relieved the labor-intensive human annotations for large-scale training data by only requiring a small number of them. The proposed method detects the elliptical regions by identifying all five key parameters of an ellipse. Experimental results showed that the proposed method achieved effective and accurate detection results in terms of both ellipse-level and bounding-box-level evaluation metrics, especially on images with degradations.

In the future, we plan to focus on multi-scale fiber detection, since deep learning based detector is not robust to detect small objects. However, the size of fibers on each image slice is relatively small. In addition, for the semi-supervised learning, we utilized synthetic data to enlarge the dataset, but if the object shape is irregular, it is hard to synthesize it. More effective semi-supervised or weakly supervised methods should be explored in the future.



\bibliographystyle{IEEEtran}
\bibliography{Fiber-TIP2015}
\begin{IEEEbiography}[{\includegraphics[clip,width=1in,height=1.25in]{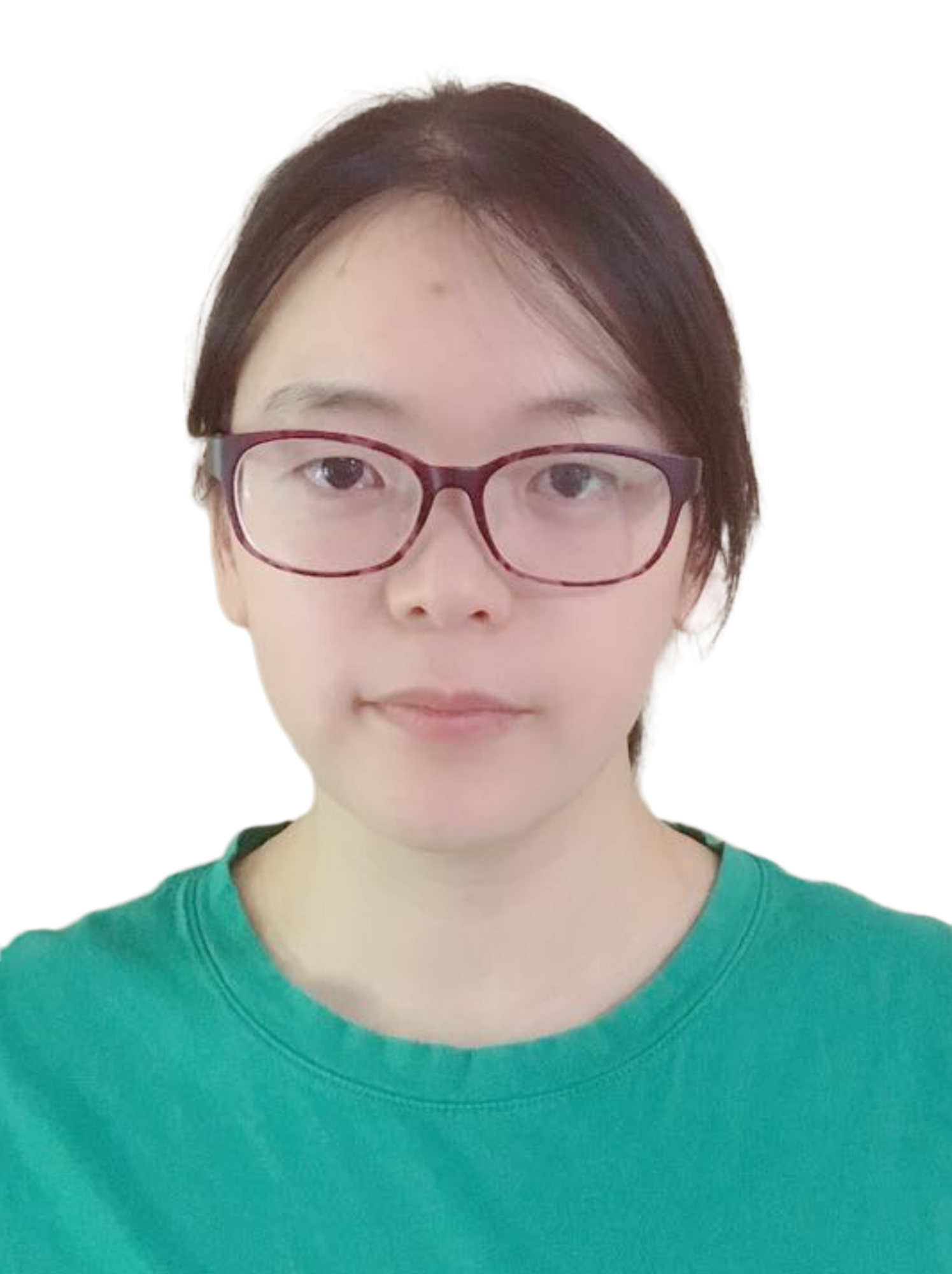}}]
{Lan Fu} received her M.S. from the Department of Biomedical Engineering, Tianjin University, Tianjin, China, in 2014. She is currently pursuing the Ph.D. degree with the Department of Computer Science and Engineering, University of South Carolina, SC, USA. Her current research interests include computer vision and machine learning. She is a student member of the IEEE.
\end{IEEEbiography}
\begin{IEEEbiography}[{\includegraphics[clip,width=1in,height=1.05in]{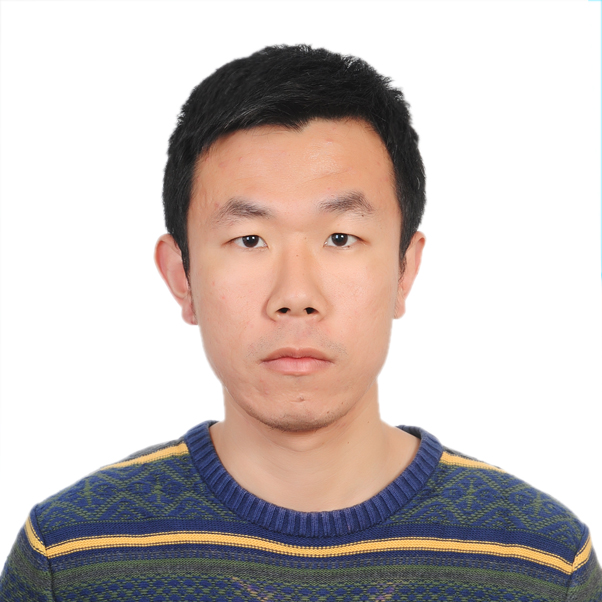}}]
{Zhiyuan Liu} received the B.S. and M.S. degrees from the Department of Biomedical Engineering, Sun Yat-sen University, Guangzhou, China, in 2010 and 2013, respectively. He is currently a Ph.D. candidate in the Department of Computer Science at University of North Carolina at Chapel Hill, NC, USA. His current research interests include computer vision, medical image analysis and statistical shape analysis.
\end{IEEEbiography}
\begin{IEEEbiography}[{\includegraphics[clip,width=1in,height=1.40in]{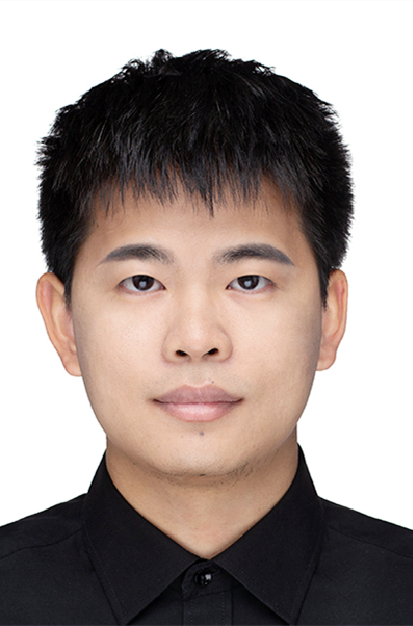}}]
{Jinlong Li} received the B.S. degree from the School of Highway, Chang’an University, Xi’an, China, in 2018. He is currently a Master student in the School of Information Engineering, Chang'an University, Xi'an, China. His major is the Computer Science and Technology. His research interests include intelligent transportation system, computer vision and deep learning.
\end{IEEEbiography}

\begin{IEEEbiography}[{\includegraphics[clip,width=1in,height=1.25in]{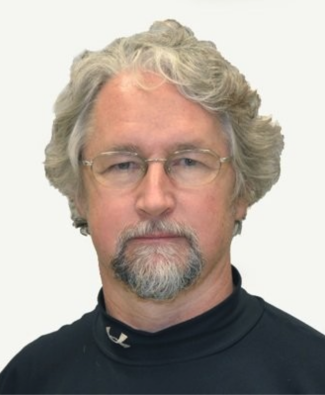}}]
{Jeff Simmons} is a materials and imaging scientist in the Metals Branch, Materials and Manufacturing Directorate of the Air Force Research Laboratory (AFRL). He received the B.S. degree in metallurgical engineering from the New Mexico Institute of Mining and Technology, Socorro, NM, USA, and M.E. and Ph.D. degrees in Metallurgical Engineering and Materials Science and Materials Science and Engineering, respectively, from Carnegie Mellon University, Pittsburgh, PA, USA. After receiving the Ph.D. degree, he began work at AFRL as a post-doctoral research contractor. In 1998, he joined AFRL as a Research Scientist. His research interests are in computational imaging for microscopy and has developed advanced algorithms for analysis of large image datasets. Other research interests have included phase field (physics-based) modeling of microstructure formation, atomistic modeling of defect properties, and computational thermodynamics. He has lead teams developing tools for digital data analysis and computer resource integration and security. He has overseen execution of research contracts on computational materials science, particularly in prediction of machining distortion, materials behavior, and thermodynamic modeling. He has published in the Materials Science, Computer Vision, Signal Processing, and Imaging Science fields. He is a member of ACM and a senior member of IEEE.
\end{IEEEbiography}
\begin{IEEEbiography}[{\includegraphics[clip,width=1in,height=1.25in]{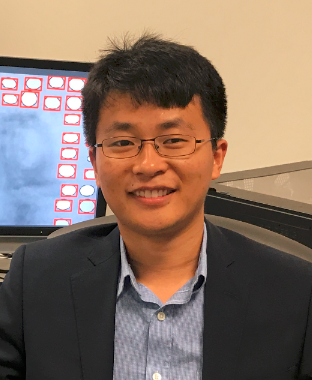}}]
{Hongkai Yu} received the Ph.D. degree in computer science and engineering from University of South Carolina, Columbia, SC, USA, in 2018. He is currently an Assistant Professor in the Department of Electrical Engineering and Computer Science at Cleveland State University, Cleveland, OH, USA. His research interests include computer vision, machine learning, deep learning and intelligent transportation systems.
\end{IEEEbiography}
\begin{IEEEbiography}[{\includegraphics[clip,width=1in,height=1.25in]{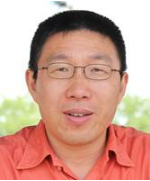}}]
{Song Wang} received the Ph.D. degree in electrical and computer engineering from the University of Illinois at Urbana–Champaign (UIUC), Champaign, IL, USA, in 2002. He was a   Research Assistant with the Image Formation and Processing Group, Beckman Institute, UIUC, from 1998 to 2002. In 2002, he joined the Department of Computer Science and Engineering, University of  South Carolina, Columbia, SC, USA, where he is currently a Professor. His current research interests include computer vision, image processing, and machine learning. Dr. Wang is a senior member of IEEE and a member of the IEEE Computer Society. He is currently serving as the Publicity/Web Portal Chair for the Technical Committee of Pattern Analysis and Machine Intelligence of the IEEE Computer Society and as an Associate Editor for the IEEE Transactions on Pattern Analysis and Machine Intelligence, Pattern Recognition Letters,and Electronics Letters.
\end{IEEEbiography}

\end{document}